%% file: main.tex
\documentclass[10pt,twocolumn,letterpaper]{article}

\usepackage[final]{cvpr}      
\usepackage{pifont}   
\usepackage{xcolor}   
\usepackage{booktabs}
\usepackage{makecell}          
\usepackage{colortbl,xcolor}   
\usepackage[normalem]{ulem}
\usepackage{subcaption}
\definecolor{rowgray}{gray}{0.92}   

\usepackage{bbm}
\input{preamble}
\definecolor{cvprblue}{rgb}{0.21,0.49,0.74}
\usepackage[pagebackref,breaklinks,colorlinks,allcolors=cvprblue]{hyperref}

\def\paperID{6163} 
\def\confName{CVPR}
\def\confYear{2026}

\title{MicroVQA++: High-Quality Microscopy Reasoning Dataset with Weakly Supervised Graphs for Multimodal Large Language Model}

\author{Manyu Li\thanks{These authors contributed equally} \\
Fudan University \\
{\tt\small 24210240029@m.fudan.edu.cn}
\and
Ruian He\textsuperscript{*} \\
Fudan University \\
{\tt\small rahe16@fudan.edu.cn}
\and 
Chenxi Ma\thanks{Corresponding Authors.} \\
Fudan University \\
{\tt\small cxma17@fudan.edu.cn}
\and 
Weimin Tan\textsuperscript{\dag} \\
Fudan University \\
{\tt\small wmtan@fudan.edu.cn}
\and 
Bo Yan\textsuperscript{\dag} \\
Fudan University \\
{\tt\small byan@fudan.edu.cn}
}

\begin{document}
\maketitle
\input{sec/0_abstract}    
\input{sec/1_intro}

\input{sec/2_related_work}
\input{sec/3_method}
\input{sec/4_experiments}

\input{sec/5_conclusion}
{
    \small
    \bibliographystyle{ieeenat_fullname}
    \bibliography{main}
}


\end{document}

%% file: sec/0_abstract.tex
\begin{abstract}

Multimodal Large Language Models are increasingly applied to biomedical imaging, yet scientific reasoning for microscopy remains limited by the scarcity of \textbf{large-scale}, \textbf{high-quality} training data. We introduce MicroVQA++, a three-stage, large-scale and high-quality microscopy VQA corpus derived from the BIOMEDICA archive. Stage one bootstraps supervision from expert-validated figure–caption pairs sourced from peer-reviewed articles. Stage two applies HiCQA-Graph, a novel \textbf{h}eterogeneous graph over \textbf{i}mages, \textbf{c}aptions, and \textbf{QA}s that fuses NLI-based textual entailment, CLIP-based vision–language alignment, and agent signals to identify and filter inconsistent samples. Stage three use MultiModal Large Language Model (MLLM) agent to generate multi-choice question (MCQ) followed by human screening. The resulting release comprises a large training split and a human-checked test split whose Bloom's level hard sample distribution exceeds MicroVQA benchmark. Our work delivers (\textbf{i}) a quality-controlled dataset that couples expert literature with graph-based filtering and human refinement; (\textbf{ii}) HiCQA-Graph, the first graph that jointly models (image, caption, QA) for cross-modal consistency filtering; (\textbf{iii}) evidence that careful data construction enables 4B-scale MLLMs to reach competitive microscopy reasoning performance (e.g., GPT-5) and achieve state-of-the-art performance among open-source MLLMs. \href{https://github.com/ieellee/MicroVQA-PlusPlus}{Code and dataset} will be released after the review process concludes.
\vspace{-2mm}

\end{abstract}

%% file: sec/1_intro.tex
\section{Introduction}

\begin{figure}[t]
  \centering
   \includegraphics[width=\linewidth]{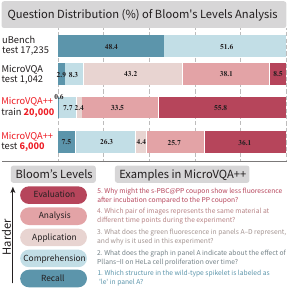}
   \caption{Bloom's levels across microscopy multimodal datasets. MicroVQA++ exhibits a substantially higher proportion in harder level and absolute count of higher-difficulty questions than MicroVQA, reflecting a stricter and more demanding evaluation setting.}
   \label{fig1}
   \vspace{-6pt}
\end{figure}

With the explosive growth of data in the modern era, many AI applications have rapidly advanced. In particular, MLLMs have shown impressive progress in domains such as industrial anomaly detection~\cite{jiang2024anomaly}, medical imaging~\cite{nam2025medicalimaging}, autonomous driving~\cite{zhou2024autonomous}, and remote sensing~\cite{zhou2025remote}. However, in the microscopy domain, most existing work on MLLMs has focused on exploring how to interface these models with microscope hardware~\cite{mandal2025evaluating, choudhary2025microscopygpt}, rather than improving their scientific reasoning ability on microscopy data itself. A key bottleneck is the \textbf{lack of high-quality multimodal microscopy datasets}, which makes it difficult to finetune MLLMs effectively.

Recently, MicroVQA~\cite{burgess2025microvqa} was introduced as the benchmark dataset targeting microscopy question answering with MLLMs, signaling an initial step toward systematic evaluation of MLLMs on microscopy tasks. Although MicroVQA is challenge enough, it contains only 1,042 samples for evaluation; it does not address the data scarcity problem that limits further progress. As MLLMs continue to scale and improve, such a small benchmark is suitable only for evaluating their reasoning performance and is unlikely to further enhance it. There is still no large-scale, high-quality multimodal microscopy dataset suitable for supervised fine-tuning of MLLMs, which severely restricts advances in microscopy-focused multimodal reasoning.


To construct such a dataset, we leverage a strong MLLM agent to generate supervision. The recent release of the BIOMEDICA~\cite{lozano2025biomedica} dataset offers a promising path forward. It is a large automatically curated biomedical vision–language dataset constructed from \textit{PubMed Central Open Access} articles, containing approximately 24M image–caption pairs. Importantly, around \textbf{10.4\%} of these samples are microscopy-related, providing strong evidence that large-scale, domain-relevant microscopy data is available. Specifically, given an image and its caption, we first extract key factual answers from the caption, and then generate questions that are grounded in the corresponding image and aligned with realistic microscopy analysis workflows. The question types follow the scientific workflow design introduced in MicroVQA, which characterizes three core experimental reasoning capabilities in microscopy. These capabilities require both abductive reasoning (inferring the most plausible explanation among multiple hypotheses) and deductive reasoning (applying general biological or experimental principles to a specific observation): \textbf{(i) Expert Visual Understanding} (\texttt{EU}): describing and interpreting visual phenomena in the image; \textbf{(ii) Hypothesis Generation} (\texttt{HG}): proposing mechanistic hypotheses that could explain the observed experimental data; \textbf{(iii) Experiment Proposal} (\texttt{EP}): suggesting follow-up experiments to validate or falsify the proposed hypotheses. After constructing the initial QA database, we further prompt the MLLM agent to convert QAs into MCQs, and ask it to explicitly provide chain-of-thought (CoT)~\cite{wei2022cot} style rationales for MCQ construction.

However, automatically generated supervision from an MLLM agent is not perfect. The agent can still introduce errors, shortcuts, or hallucinations. To systematically improve data quality, we introduce a \textbf{H}eterogeneous \textbf{I}mage-\textbf{C}aption-\textbf{QA} Graph (HiCQA-Graph) that explicitly models the relationships between images, their captions, and the generated QAs. Intuitively, our goal is to capture cross-modal consistency so that we can identify and filter out low-quality or hallucinated samples. To the best of our knowledge, \emph{MicroVQA++} is the first large-scale multimodal VQA training dataset specifically tailored to microscopy. After supervised fine-tuning (SFT) on MicroVQA++ training data, an MLLM with only 4B parameters matches the reasoning performance of the strongest proprietary commercial model GPT-5 on the MicroVQA benchmark, and establishes new state-of-the-art results among open-source MLLMs. Our main contributions are as follows:



\begin{itemize}
    \item \textbf{A novel pipeline for QA and MCQ construction.} Starting from high-quality image–caption pairs extracted from the literature, we use a powerful MLLM agent to generate grounded QA pairs, and then derive multiple-choice questions with explicit rationales.
    \item \textbf{HiCQA-Graph for cross-modal consistency filtering.} We propose a Heterogeneous Image-Caption-QA Graph that jointly models images, captions, and QAs. By leveraging Natural Language Inference (NLI) based entailment ~\cite{bowman2015large} between caption and QA, CLIP-based vision–language alignment between image and QA, and agent-derived supervision signals, our graph learns to identify and filter unreliable samples, thus improving dataset quality.
    \item \textbf{MicroVQA++ and strong microscopy reasoning at 4B scale.} With finetuning on the MicroVQA++ training set, an MLLM with only 4B parameters already on par with GPT-5 on the MicroVQA benchmark and achieves new state-of-the-art performance among open-source MLLMs in microscopy reasoning.
\end{itemize}

\begin{table}[t]
    \centering 
    \small
    \caption{Compared to MicroVQA, Ours contains 19.2× and 5.8× more questions in the training and test splits, respectively, and 33.7× and 20.4× more images in the corresponding splits, demonstrating a substantially larger scale and richer visual coverage.} 
    \vspace{-6pt}
    \label{tab1}
    \begin{tabular}{llll}
    \Xhline{1.2pt}
    Dataset feature & MicroVQA & Ours\textsuperscript{train} & Ours\textsuperscript{test} \\
    \hline
    Total Qs & 1042 & 20000\textsubscript{19.2$\times$} & 6000\textsubscript{5.8$\times$} \\
    Unique images & 255 & 8594\textsubscript{33.7$\times$} & 5198\textsubscript{20.4$\times$} \\ 
    CoTs & 0 & 20000 & 6000 \\
    Avg. MCQ Q len & 66 & 162 & 123 \\
    Avg. MCQ A len & 15 & 79 & 55 \\
    Image Modalities & \multicolumn{3}{r}{Light, Fluoro, Electron} \\
    Image Scales & \multicolumn{3}{r}{Tissue, Cell, Subcell, Atomic} \\
    \Xhline{1.2pt}
\end{tabular}
\vspace{-5pt}
\end{table}

%% file: sec/2_related_work.tex
\section{Related work}





\begin{figure*}[t]
    \centering
    \includegraphics[width=\linewidth]{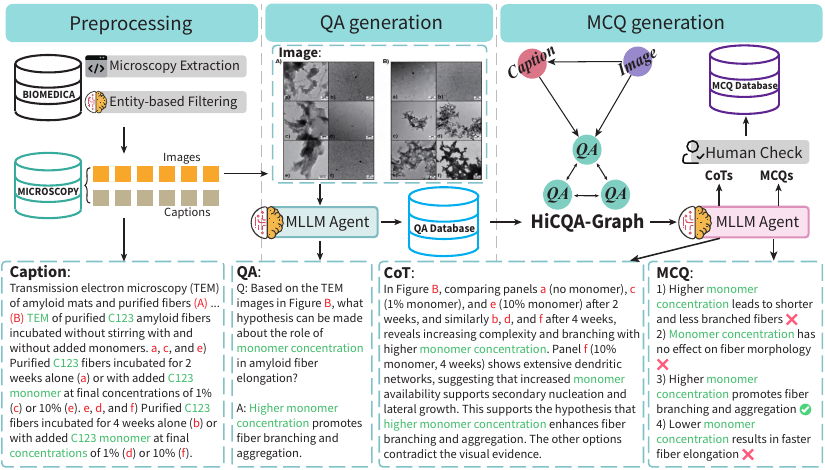}
    \caption{MicroVQA++ is built in three stages. We first sample figure–caption pairs from the Microscopy category of BIOMEDICA. An MLLM agent then extracts answer spans from the captions to construct initial QA pairs, leveraging peer-reviewed articles to ensure expert-validated supervision. Next, we pass the data through HiCQA-Graph, which evaluates cross-modal consistency to judge generation quality. Finally, conditioned on the validated items (human check pipeline available in appendix), an MLLM agent produces CoT rationales and MCQ variants for each question. \textcolor{red}{red} indicates grounding informations and \textcolor{green!70!black}{green} indicates important entities.} 
    \label{fig2}
    \vspace{-12pt}
\end{figure*}

\subsection{MLLM benchmarks in microscopy}
Benchmarks for evaluating MLLM reasoning in microscopy remain limited. Figure~\ref{fig1} shows proportional distribution of Bloom’s level~\cite{goals1956bloom}, used to measure question difficulty in the dataset. Higher Bloom's levels are more cognitively challenging. \emph{$\mu$Bench}~\cite{lozano2024ubench} assembles a broad microscopy understanding suite with 22 tasks and 17{,}235 images, but its scientific depth is shallow. \emph{MicroVQA} abstracts microscopy research workflows into three capacities and emphasizes deeper reasoning, yet contains only 255 images and 1{,}042 QA pairs. Building on high-quality figure–caption pairs from scientific articles, HiCQA-Graph filtering, and MLLM-agent generation, we construct \emph{MicroVQA++} with a 20K-question train set and a 6K-question test set, far larger than MicroVQA and with a substantially higher proportion of depth-oriented items. Tabel~\ref{tab1} shows dataset features among \emph{MicroVQA} and \emph{MicroVQA++}.

\subsection{QA and MCQ construction}
Generating multiple-choice questions with strong distractors is critical for human education and machine learning evaluation~\cite{gierl2017mcq1, alhazmi2024mcq2, bitew2023qagen1}. Existing pipelines often rely on prompt engineering and expert-heavy designs: \emph{$\mu$Bench} converts collected image–QA pairs to MCQs using GPT-4o before expert verification, while \emph{MicroVQA} adopts a two-stage expert–agent strategy. These approaches, however, require complex manual heuristics and underuse the inherent ties among images, captions, and QAs. Our proposed \emph{HiCQA-Graph} explicitly models \texttt{Image}, \texttt{Caption} and \texttt{QA} node types and learns their relationships under weak supervision, enabling principled filtering that yields new, higher-quality QA/MCQ data with CoT.

\subsection{Graph-based data generation and filtering}
Graph methods are widely used to construct or denoise training sets from weak supervision or noisy labels ~\cite{boykov2001g1, krahenbuhl2011g2, ahn2018g3}. A common pipeline fuses multi-source weak signals into a “label graph,” followed by graph-based denoising and consolidation~\cite{ratner2016relatedgraph1}. Label propagation~\cite{zhu2002relatedgraph2} provides a classic framework for diffusing sparse, reliable seeds over a similarity graph. In visual weak supervision and noisy-label learning, DualGraph~\cite{zhang2021dualgraph} jointly reasons over a sample graph and a label graph to infer noise, while NGC~\cite{wu2021ngc} exploits inter-sample geometry to select clean instances, both embodying the principle that structural consistency implies sample confidence. In this work, we treat a graph as an intermediate for data filtering and amplify multi-source consistency signals (e.g., image–text alignment, textual entailment) via message passing with GraphSAGE~\cite{hamilton2017graphsage} and GAT~\cite{velickovic2017gat}.

%% file: sec/3_method.tex
\section{Method}

\subsection{Motivation}
A persistent bottleneck in microscopy is the lack of high-quality multimodal datasets, which substantially limits the field’s potential. The emergence of BIOMEDICA offers a promising path forward. This is valuable for two reasons: (\textbf{i}) Because the images and captions are extracted from scientific publications, the captions are already written, checked, and curated by human experts, leading to comparatively high semantic quality~\cite{baghbanzadeh2025moti1, ruckert2024moti2, lin2023moti3, zhang2023moti4, li2023llavamed}. (\textbf{ii}) The microscopy subset alone accounts for roughly 2.5M image–caption pairs, which forms a solid foundation for building a large-scale multimodal microscopy dataset.


\subsection{HiCQA-Graph}
\subsubsection{Notations formulation}
In this paper, we consider a HiCQA-Graph $\mathcal{G} = (\mathcal{V}, \mathcal{E}, \mathcal{I}, \mathcal{T})$ with $|\mathcal{V}| = n$ nodes, $|\mathcal{E}| = m$ edges, $|\mathcal{I}| + |\mathcal{T}| = n$ sets of image and text. Each node $v_i \in \mathcal{V}$ is associated with an embedded image $\textbf{v} \in \mathcal{I}$ or embedded text $\textbf{t} \in \mathcal{T}$. 

\subsubsection{Node definition}
\label{sec:dataset_nodes}





\begin{figure}[t]
  \centering
   \includegraphics[width=\linewidth]{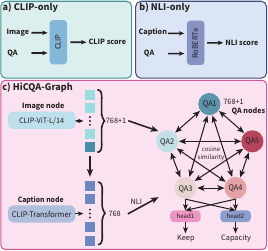}
   \caption{\textbf{a)} and \textbf{b)} shows CLIP and NLI only filtering method. \textbf{c)} indicates HiCQA-Graph structure. Cross-modal consistency token is added to Image and QA nodes. Two heads are used to predict soft weak supervised labels.}
   \label{fig3}
   \vspace{-6mm}
\end{figure}

Figure~\ref{fig3} illustrates the overall architecture of HiCQA-Graph, comprising three node types: \texttt{Image}, \texttt{Caption}, and \texttt{QA}.
\paragraph{Image node.}
We encode each image with \textbf{CLIP ViT-L/14}~\cite{radford2021clip, dosovitskiy2020vit}. Before encoding, all images are resized to $224\times224$. Let the CLIP visual embedding be $\mathbf{v}_i\in\mathbb{R}^{f}$, caption text embedding be $\mathbf{t}_i\in\mathbb{R}^{f}$ (here, $f{=}768$). We further append a one-dimensional \emph{consistency score} between the image and its caption:\footnote{We L2-normalize CLIP embeddings and use cosine similarity unless otherwise stated.}
\begin{equation}
\begin{cases}
c^{\text{img-cap}}_{i} ~=~ \operatorname{cos}\!\big(\mathbf{v}_i, \mathbf{t}_i\big)\in[-1,1] \\
\tilde{c}^{\text{img-cap}}_{i} ~=~ \frac{c^{\text{img-cap}}_{i}+1}{2}\in[0,1].
\end{cases}
\end{equation}
The final image node feature is the concatenation $\mathbf{x}^{\text{img}}_i=[\,\mathbf{v}_i;\tilde{c}^{\text{img-cap}}_{i}\,]\in\mathbb{R}^{f+1}$. Each image node has a single outgoing edge to its unique caption node and outgoing edges to all associated QA nodes.

\paragraph{Caption node.}
We encode the caption with the \textbf{CLIP text Transformer}. Due to CLIP's design, the maximum effective text length is 75 tokens (excluding BOS/EOS). To avoid truncation, overlong captions are first summarized to a suitable length by an MLLM-based agent and then fed to CLIP, yielding $\mathbf{x}_{i}^{cap} \in \mathbb{R}^{f}$. Each caption node connects to all QA nodes of the same sample with edge attributes given by NLI: for caption $\rightarrow$ answer, we take the entailment probability:
\begin{equation}
p^{\text{ent}}_{i,k} \in [0,1]
\end{equation}
as the edge attribute on $(\texttt{Caption}\!\rightarrow\!\texttt{QA}_{k})$ because the answer should be supported by the caption.
\paragraph{QA node.}
For each QA pair, we concatenate question and answer texts, apply the same text preprocessing and CLIP encoding to obtain $\mathbf{q}_{i,k}\in\mathbb{R}^{f}$. We also compute an image--QA consistency score
\begin{equation}
\begin{cases}
c^{\text{img-qa}}_{i,k} ~=~ \operatorname{cos}\!\big(\mathbf{v}_i, \mathbf{q}_{i,k}\big) \in [-1,1] \\
\tilde{c}^{\text{img-qa}}_{i,k} ~=~ \frac{c^{\text{img-qa}}_{i,k}+1}{2}\in[0,1]
\end{cases}
\end{equation}
and form the \texttt{QA} node feature $\mathbf{x}^{\text{qa}}_{i,k}=[\,\mathbf{q}_{i,k};\tilde{c}^{\text{img-qa}}_{i,k}\,]\in\mathbb{R}^{f+1}$. Within each sample, QA nodes are fully connected with \emph{directed} edges in both directions excluding self-loops; the edge attribute on $(\texttt{QA}_{k}\!\rightarrow\!\texttt{QA}_{\ell})$ is the nonnegative cosine similarity
\begin{equation}
a^{\text{qa-qa}}_{i,k\to\ell} ~=~ \max\!\bigl\{0,~\operatorname{cos}(\mathbf{q}_{i,k},\,\mathbf{q}_{i,\ell})\bigr\}.
\end{equation}

\paragraph{Weak labels for QA nodes.}
We define two supervision signals on \texttt{QA} nodes:
\begin{itemize}
\item \textbf{(i) Keep score (weak supervised).}
We combine the image--QA CLIP similarity with the \texttt{Caption}$\rightarrow$\texttt{Answer} NLI entailment probability into a scalar confidence and obtain soft \texttt{Keep} label. A simple and effective fusion is the weighted sum
\begin{equation}
\label{eq:keep-score}
y^{\text{keep}}_{i,k} ~=~ \alpha \cdot \tilde{c}^{\text{img-qa}}_{i,k} ~+~ (1{-}\alpha)\cdot p^{\text{ent}}_{i,k}, 
\ \alpha\in[0,1],
\end{equation}

\item \textbf{(ii) Capacity (supervised).}
We use a 3-way label
\begin{equation}
y^{\text{cap}}_{i,k}\in\{\text{EU},~\text{HG},~\text{EP}\},
\end{equation}
indicating capacity generated by MLLM-based agent.
\end{itemize}

\subsubsection{Graph neural network}
\label{sec:gnn}

We define four directed relation types $\mathcal{R}=\{ \textit{describe\_by, \ asked\_about, \ supports, \ similar} \}$. Edges are typed as (\texttt{src}, \textit{relation}, \texttt{dst}): (\texttt{Image}, \textit{described\_by}, \texttt{Caption}), (\texttt{Image}, \textit{asked\_about}, \texttt{QA}), (\texttt{Caption}, \textit{supports}, \texttt{QA}), and (\texttt{QA}, \textit{similar}, \texttt{QA}). Following Sec.~\ref{sec:dataset_nodes}, node features are CLIP embeddings with an additional 1-D consistency score for \textit{Image} and \textit{QA}, and plain CLIP text features for \textit{Caption}.

\paragraph{Input projections.}
Let $\mathbf{x}^{\text{img}}\!\in\!\mathbb{R}^{f+1}$, 
$\mathbf{x}^{\text{cap}}\!\in\!\mathbb{R}^{f}$, 
$\mathbf{x}^{\text{qa}}\!\in\!\mathbb{R}^{f+1}$ be node features. 
We project them into a common hidden space $\mathbb{R}^{d}$ via separate linear layers. 
\paragraph{Heterogeneous message passing.} We stack $L$ hetero-convolution layers (\texttt{HeteroConv}~\cite{fey2019fast}, $L{=}2$ default) with \emph{sum} aggregation over relations. For each layer $\ell$, we maintain per-relation operators and sum their outputs for every node type.

\paragraph{SAGEConv on cross-type edges.}
For (\texttt{Image}, \textit{described\_by}, \texttt{Caption}) and 
(\texttt{Image}, \textit{asked\_about}, \texttt{QA}), 
we use \textbf{GraphSAGE}~\cite{hamilton2017graphsage} without edge attributes.
Given a destination node $u$ (\texttt{Caption} or \texttt{QA}), let $\mathcal{N}_\ell(u)$ be the set of incoming neighbors of type \texttt{Image} at layer $\ell$. A standard SAGE update is

\begin{equation}
\label{eq:sage}
\begin{aligned}
\hat{\mathbf{h}}^{(\ell+1)}_u
~=~ \sigma\!\Big(
&\mathbf{W}^{(\ell)}_{\text{self}} \mathbf{h}^{(\ell)}_u
\\
&+ \mathbf{W}^{(\ell)}_{\text{neigh}}\,
  \operatorname{AGG}\{\mathbf{h}^{(\ell)}_v: v\in\mathcal{N}_\ell(u)\}
\Big)
\end{aligned}
\end{equation}

where $\operatorname{AGG}$ is usually mean aggregation and $\sigma$ is a nonlinearity ReLU.

\paragraph{GATv2Conv with edge attributes.}
For (\texttt{Caption}, \textit{supports}, \texttt{QA}) and 
(\texttt{QA}, \textit{similar}, \texttt{QA}), 
we adopt \textbf{GATv2}~\cite{brody2021gatv2} with $4$ attention heads and \emph{edge attributes} (edge\_dim$=1$). 
Let $e_{uv}\!\in\!\mathbb{R}$ be the scalar edge attribute: NLI entailment probability for \textsc{supports} and nonnegative cosine similarity for \textsc{similar}. 
GATv2 computes attention coefficients using a content-based mechanism:
\begin{equation}
\begin{aligned}
\alpha^{(\ell)}_{uv}
~=~ \operatorname{softmax}_{v\in\mathcal{N}_\ell(u)} \big(\mathbf{a}^{\top}\, \phi\!\Big(\mathbf{W}\mathbf{h}^{(\ell)}_u \Vert~ \mathbf{W}\mathbf{h}^{(\ell)}_v 
\\
\Vert~ \mathbf{w}_e\, e_{uv}
\Big)\big),
\end{aligned}
\end{equation}

and aggregates messages as
\begin{equation}
\label{eq:gat}
\hat{\mathbf{h}}^{(\ell+1)}_u
~=~
\sigma\!\Big(
\sum\nolimits_{v\in\mathcal{N}_\ell(u)}
\alpha^{(\ell)}_{uv}\,\mathbf{W}\mathbf{h}^{(\ell)}_v
\Big),
\end{equation}
where $\phi$ is a pointwise nonlinearity (e.g., LeakyReLU), $\Vert$ denotes concatenation, and multi-head outputs are averaged to match the hidden size $d$.

\paragraph{Relation-wise fusion.}
For each node type, we sum the relation-specific outputs:
\begin{equation}
\tilde{\mathbf{h}}^{(\ell+1)}_u
~=~
\sum\nolimits_{r\in\mathcal{R}(u)}
\hat{\mathbf{h}}^{(\ell+1)}_{u,r},
\end{equation}
where $\mathcal{R}(u)$ are relations incident to the type of $u$ in layer $\ell$.
We then apply residual connection, LayerNorm, and dropout:
\begin{equation}
\mathbf{h}^{(\ell+1)}_u
~=~
\operatorname{Dropout}\!\Big(
\operatorname{LN}\big(\sigma(\tilde{\mathbf{h}}^{(\ell+1)}_u) + \mathbf{h}^{(\ell)}_u\big)
\Big).
\end{equation}

\vspace{2pt}\noindent\textbf{Task heads on QA nodes.}
Only \texttt{QA} nodes are supervised. Given the final representation $\mathbf{h}^{(L)}_{\text{qa}}$, we use two MLP heads:
\begin{equation}
\begin{cases}
\mathbf{z}^{\text{keep}} &= \mathrm{MLP}_{\text{keep}}(\mathbf{h}^{(L)}_{\text{qa}})\in\mathbb{R}^{2}, 
\\
\mathbf{z}^{\text{cap}} &= \mathrm{MLP}_{\text{cap}}(\mathbf{h}^{(L)}_{\text{qa}})\in\mathbb{R}^{3},
\end{cases}
\end{equation}
producing soft logits for \textsc{Keep} and 3-way \textsc{EU/HG/EP}, respectively. 
In training, we optimize a multi-task objective (cross-entropy for both heads) with optional class weights; gradients are clipped and dropout is applied after each layer as in the implementation.

Intuitively, \text{SAGEConv} propagates \emph{cross-modal} evidence from \texttt{Image} $ \!\rightarrow\!$ \texttt{Caption}/\texttt{QA} via mean-aggregated neighborhood messages \eqref{eq:sage}, while \text{GATv2Conv} leverages \emph{edge-aware} attention \eqref{eq:gat} to weight \texttt{Caption} $ \!\rightarrow\!$ \texttt{QA} by NLI entailment and \texttt{QA} $ \!\leftrightarrow\!$ \texttt{QA} by textual similarity. Relation-wise summation and per-type normalization stabilize heterogeneous fusion across the HiCQA-Graph.


\begin{figure*}[t]
    \centering
    \includegraphics[width=\linewidth]{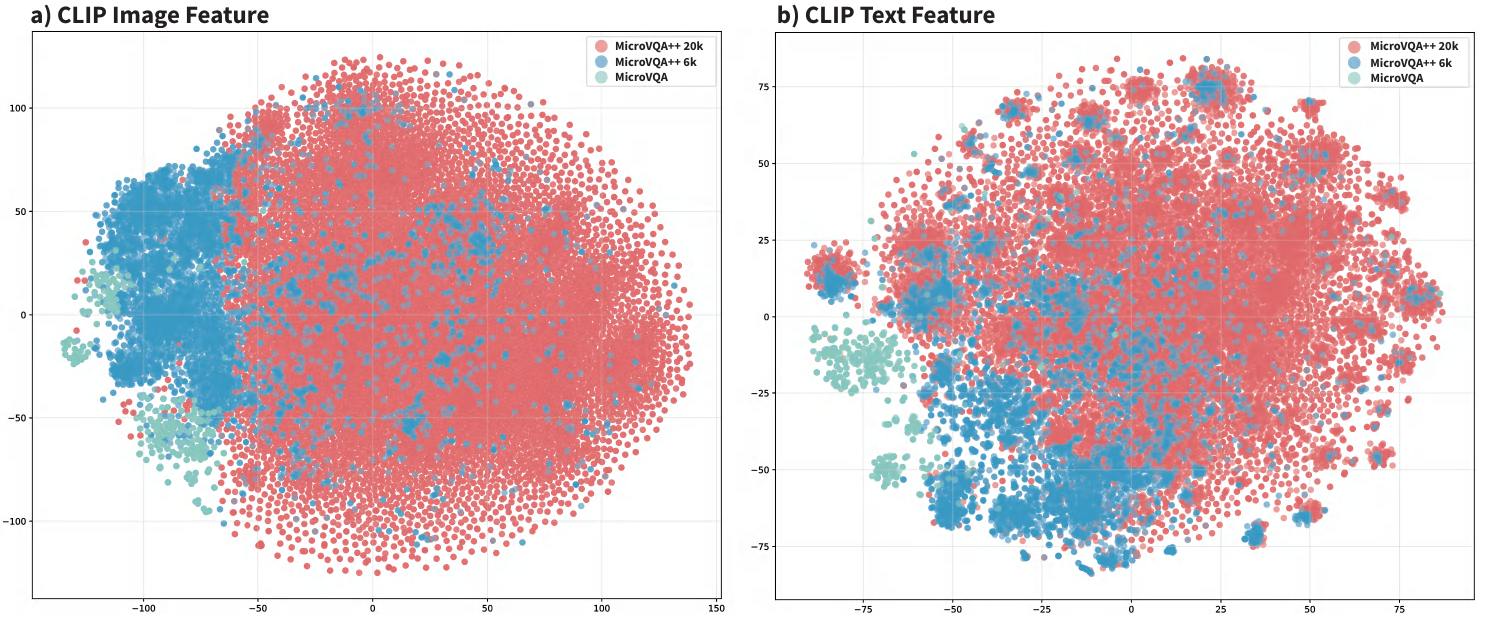}
    \caption{Two-dimensional t-SNE of CLIP embeddings for images and questions. t-SNE uses perplexity of 30 and 1,000 iterations.}
    \label{fig4}
    \vspace{-12pt}
\end{figure*}

\subsection{Data generation}
\subsubsection{QA generation pipeline}

Deriving answers directly from the captions in BIOMEDICA yields a subset of data that is generally high quality. Based on each figure and its associated answer, we further employ an MLLM agent to generate QA samples spanning three capacity types: \texttt{EU}, \texttt{HG}, and \texttt{EP}. Figure~\ref{fig2} shows our generation pipeline.

When constructing HiCQA-Graph, we are constrained by the maximum number of text tokens. For QA pairs whose length exceeds this limit, we apply summarization to compress the content while preserving most of the original information, more details in appendix. 

\subsubsection{MCQs generation pipeline}




Once a high-quality QA database is obtained, we employ an MLLM agent to generate three high-quality distractor options and an accompanying explanation (used as CoT). For the MicroVQA++ test set, we subsequently conduct a human review to identify any significant errors. MicroVQA++ exhibits no information leakage with respect to MicroVQA. Detailed dataset licensing information is provided in the appendix. 





\subsection{Data quality analysis}

\subsubsection{Bloom's level}
Figure~\ref{fig1} presents the Bloom’s levels of our dataset. The evaluation method follows \cite{goals1956bloom}, which defines question difficulty. We follow \emph{MicroVQA} and use an LLM to assess the difficulty of questions in our dataset. As shown, our dataset contains a substantially higher proportion of high-difficulty samples and a larger overall scale than \emph{MicroVQA}.
\subsubsection{CLIP feature distribution}
Figure~\ref{fig4} highlights dataset-specific geometry in the shared CLIP space using t-SNE~\cite{maaten2008visualizing}.
\paragraph{Intra-dataset compactness varies.} \textbf{(i)} Image feature space. MicroVQA++ exhibits a compact and uniform arrangement, whereas MicroVQA forms two relatively tight clusters, indicating greater heterogeneity in its image distribution. MicroVQA++ also covers a broader range of images. \textbf{(ii)} Text feature space. MicroVQA++ and MicroVQA present clear distributional boundaries, highlighting substantial differences between the questioning styles and perspectives of human experts and those of MLLM agents.
\paragraph{Inter-dataset overlap is non-negligible.} Overlap regions indicate transfer-friendly zones where content and language conventions align; these are the regimes where cross-dataset generalization after SFT tends to be strongest. 
\paragraph{Cross-modal co-location.} Image points and Question points from the same dataset lie in nearby neighborhoods, indicating that CLIP’s alignment is already a strong prior; this justifies using CLIP-based signals inside our graph filter and helps interpret why combining textual NLI with vision–language alignment is superior to either alone.

\subsection{MLLMs training details}

With the MicroVQA++ training split in place, we explored both the data format (free-form QA vs. close-form MCQ) and the learning algorithm (Supervised Fine-Tuning, SFT, vs. Group Relative Policy Optimization, GRPO). Unless otherwise noted, training used an InternVL-style instruct model; microscopy images were resized so the short side was 448 pixels.

\subsubsection{Supervised fine-tuning}
We fine-tuned with LLaMA-Factory~\cite{zheng2024llamafactory}, using LoRA~\cite{hu2022lora} with rank 16 applied to all attention projections (q, k, v, o) and MLP blocks (gate, up, down). For optimizer, we use AdamW with a 1e-4 learning rate, weight decay 0.1, cosine decay with 5\% warm-up; 3 epochs in bfloat16. We enabled gradient checkpointing and gradient clipping at 1.0. To stabilize throughput, we used sequence packing (context length 4,096) and gradient accumulation to reach an effective batch size of 256 sequences.

\paragraph{QA-based SFT.} For free-form \texttt{QA} samples, the input bundled a system instruction, one image, and a question; the target was the full textual answer. A microscopy-oriented instruction encouraged concise, evidence-grounded responses. We optimized next-token cross-entropy over the entire answer span. In ablations, QA-only supervision improved factual grounding and descriptive precision but did not fully train option discrimination.

\paragraph{MCQ-based SFT.} For \texttt{MCQ} samples, the input included the system instruction, an image, a question, and four options. The model was trained to produce a brief chain-of-thought rationale and a structured final choice tag. 

\subsubsection{Group relative policy optimization}

We trained GRPO~\cite{grpo} algorithm with the verl~\cite{sheng2025verl} framework. For each prompt we sampled $n=6$ candidate completions, computed per-sample rewards, and shaped advantages relative to the group mean using a frozen SFT policy as reference.

\paragraph{Rewards.} We combined (i) a format reward that penalizes extraneous markup, and (ii) an answer-accuracy reward. Invalidly formatted outputs receive a small negative reward.

\paragraph{Optimization details.} We use four RTX 3090. Learning rate 5e-6, KL coefficient $\beta=0.05$, entropy bonus 0.001, value-loss coefficient 0.5, and advantage normalization per batch. We trained for 1 epochs over the MCQ subset, reusing the SFT tokenizer and image encoder. GRPO sharpened option calibration and adherence to output format; used after SFT it provided reliable incremental gains.

%% file: sec/4_experiments.tex
\section{Experiments}

\subsection{Dataset and evaluation metric}



\paragraph{MicroVQA and MicroVQA++.} In subsequent experiments, we train on the MicroVQA++ training set and evaluate on the MicroVQA and MicroVQA++ test sets. To ensure fairness, we use an identical prompt template for inference on both test sets. HiCQA-Graph training details are available in appendix. 
\vspace{-2mm}
\paragraph{Evaluation metric.} We follow the evaluation protocol of MicroVQA: for each of the three capacities (\texttt{EU}, \texttt{HG}, \texttt{EP}), we compute \texttt{MCQ} accuracy via regex-based matching.

\begin{table}[t]
    \centering
    \small
    \caption{Comparison experiments on MicroVQA. \textsuperscript{*}denotes results obtained from the MicroVQA leaderboard; \textsuperscript{\dag}denotes results after SFT on the MicroVQA++ training set. Bold mark means the \textbf{best} performance; a single underline marks the \uline{second-best}. Unless otherwise specified, the same conventions apply hereafter. The latest version of the closed-source models used are provided in the appendix.} 
    \vspace{-4pt}
    \begin{tabular}{lcccc}
    \Xhline{1.2pt}
    Models & EU & HG & EP & Avg \\
    \Xhline{0.5pt}
    \addlinespace[1.2pt]
    \Xhline{0.5pt}
    \multicolumn{5}{l}{\textbf{Baseline}}   \\ 
    Random\textsuperscript{*}~\cite{burgess2025microvqa} & 21.9 & 21.8 & 21.9 & 22.0 \\
    Human\textsuperscript{*}~\cite{burgess2025microvqa} & 52.7 & 47.5 & 51.4 & 50.3 \\
    \hline
    \multicolumn{5}{l}{\textbf{Close sourced (commercial)}}   \\ 
    GPT-4o\textsuperscript{*}~\cite{openai_gpt4o} & 48.7 & 43.1 & 44.8 & 45.6 \\
    GPT-4o-mini\textsuperscript{*}~\cite{openai_gpt4o} & 48.5 & 43.6 & 47.0 & 46.2 \\
    o1\textsuperscript{*}~\cite{openai_o1_series} & 55.4 & 50.2 & 53.0 & 52.8 \\
    Claude Sonnet 4.5\textsuperscript{*}~\cite{anthropic_claude_sonnet_4_5_system_card_2025} & 55.1 & 56.4 & 49.6 & 54.4 \\
    o4-mini\textsuperscript{*}~\cite{openai_o3_o4mini_systemcard_2025} & 57.9 & 56.1 & 50.4 & 55.6 \\
    o3\textsuperscript{*}~\cite{openai_o3_o4mini_systemcard_2025} & \uline{61.5} & \textbf{60.5} & 53.5 & \uline{59.3} \\
    GPT-5\textsuperscript{*}~\cite{openai_gpt5_system_card_2025} & \textbf{63.3} & \uline{58.9} & 53.9 & \textbf{59.4} \\
    \hline
    \multicolumn{5}{l}{\textbf{Open sourced}} \\
    LLaVA-Med-Mistral-7B\textsuperscript{*}~\cite{li2023llavamed} & 37.3 & 47.1 & 41.6 & 43.0 \\
    Qwen-2-VL-7b\textsuperscript{*}~\cite{wang2024qwen2vl} & 54.1 & 43.3 & 49.6 & 48.8 \\
    InternVL3.5-2B-Instruct~\cite{wang2025internvl35} & 46.4 & 41.0 & 47.4 & 44.4 \\
    InternVL3.5-4B-Instruct & 45.9 & 43.8 & 52.6 & 46.6 \\
    \hline
    \multicolumn{5}{l}{\textbf{MicroVQA++ train set fine-tuned}} \\
    \rowcolor{rowgray}
    LLaVA-Med-Mistral-7B\textsuperscript{\dag} & 54.6 & 51.7 & 56.1 & 53.7 \\
    \rowcolor{rowgray}
    InternVL3.5-2B-Instruct\textsuperscript{\dag} & 55.4 & 52.4 & \uline{57.0} & 54.5 \\
    \rowcolor{rowgray}
    InternVL3.5-4B-Instruct\textsuperscript{\dag} & 61.0 & 56.9 & \textbf{61.3} & \textbf{59.4} \\
    \Xhline{1.2pt}
    \end{tabular}
    \vspace{-4pt}
    \label{tab2}
\end{table}

\begin{table}[t]
    \centering
    \small
    \caption{Comparison experiments on MicroVQA++ testing set.}
    \vspace{-4pt}
    \begin{tabular}{lcccc}
    \Xhline{1.2pt}
    Models & EU & HG & EP & Avg \\
    \Xhline{0.5pt}
    \addlinespace[1.2pt]
    \Xhline{0.5pt}
    \multicolumn{5}{l}{\textbf{Baseline}} \\ 
    Random & 25.0 & 25.0 & 25.0 & 25.0 \\
    \hline
    \multicolumn{5}{l}{\textbf{Close sourced (commercial)}} \\
    GPT-4o-mini~\cite{openai_gpt4o} & 31.5 & 53.4 & \uline{42.2} & 37.3 \\
    qwen-vl-flash~\cite{bai2023qwen} & 26.1 & 49.2 & 37.6 & 32.2 \\
    \hline
    \multicolumn{5}{l}{\textbf{Open sourced}} \\
    LLaVA-Med-Mistral-7B~\cite{li2023llavamed} & 26.6 & 41.0 & 41.2 & 33.5 \\
    Qwen2.5-VL-7B-Instruct~\cite{bai2025qwen25} & 32.9 & 48.5 & 34.9 & 36.3 \\
    Qwen3-VL-4B-Instruct~\cite{yang2025qwen3} & 32.1 & 48.9 & 39.1 & 36.4 \\
    \hline
    \multicolumn{5}{l}{\textbf{MicroVQA++ train set fine-tuned}} \\
    \rowcolor{rowgray}
    LLaVA-Med-Mistral-7B\textsuperscript{\dag} & \textbf{39.3} & \textbf{61.4} & 39.3 & \textbf{45.3} \\
    \rowcolor{rowgray}
    InternVL3.5-4B-Instruct\textsuperscript{\dag} & \uline{36.1} & \uline{59.7} & \textbf{42.7} & \uline{41.3} \\
    \Xhline{1.2pt}
    \end{tabular}
    \vspace{-8pt}
    \label{tab3}
\end{table}

\subsection{Comparison experiments}



\paragraph{MicroVQA benchmark.} We evaluate both proprietary commercial systems and open-source MLLMs on the MicroVQA benchmark in Table~\ref{tab2}. Consistent with observations in the original MicroVQA paper, we find: \textbf{(i)} Small-scale MLLMs can be surprisingly competitive. For example, the base model InternVL3.5-4B-Instruct achieves average accuracy on par with GPT-4o series. \textbf{(ii)} Models that excel on natural-image multimodal data (e.g., o3 and GPT-5) still struggle with microscopy-centric reasoning that demands domain knowledge and multi-step deduction. \textbf{(iii)} SFT on our MicroVQA++ train split substantially boosts microscopy reasoning for small models: InternVL3.5-2B/4B-Instruct improve by 22.7\% and 27.5\% relatively on Avg, respectively. \textbf{(iv)} After SFT on MicroVQA++, a 4B-parameter model can match the strongest commercial models on MicroVQA despite being an order of magnitude smaller.


\paragraph{MicroVQA++ test set.} We further compare proprietary and open-source MLLMs on the more challenging \emph{MicroVQA++} test set in Table~\ref{tab3}. As illustrated in Figure~\ref{fig1}, it adopts a higher Bloom’s level harder question distribution. Since the \emph{MicroVQA++} train and test distributions have intentionally low overlap (Figure~\ref{fig4}), SFT on the train split yields only a modest +0.6 absolute gain for InternVL3.5-4B-Instruct on the test set, underscoring the reduced dataset bias and stronger generalization requirements.

\paragraph{Filtering methods.} For NLI, we fine-tune an XLM-R model~\cite{liu2019roberta, conneau2020xlmr} on XNLI and MultiNLI~\cite{conneau2018xnli, williams2018mnli} to score \texttt{caption} $\rightarrow$ \texttt{Answer} entailment. For CLIP, we use OpenCLIP~\cite{openclip} (ViT-L/14 image encoder + Transformer text encoder~\cite{vaswani2017transformer}) initialized from BIOMEDICA-pretrained weights. NCLIP is a linear combination of the NLI and CLIP scores. As shown in Table~\ref{tab4}, using only NLI or CLIP is inferior, while NCLIP improves robustness and generalization on InternVL3.5-2B/4B. Our HiCQA-Graph jointly models \texttt{Image}, \texttt{Caption}, and \texttt{QA} via heterogeneous message passing, achieving the best average performance at the top-75\% keep ratio.

\begin{table}[t]
    \centering
    \small
    \caption{Comparison experiments on data filtering methods (2B/4B). Superscript in methods indicates the subset with the top-x\% of the dataset retained. }
    \vspace{-4pt}
    \begin{tabular}{lcccc}
    \Xhline{1.2pt}
    Methods & EU & HG & EP & Avg  \\
    \Xhline{0.5pt}
    \addlinespace[1.2pt]
    \Xhline{0.5pt}

    Full\textsuperscript{100\%} & \uline{55.9}/60.0 & \uline{51.9}/55.0 & 55.7/60.9 & \uline{54.2}/\uline{58.2} \\
    \hline
    NLI\textsuperscript{25\%} & 55.1/58.7 & 49.5/54.3 & 57.0/\textbf{62.6} & 53.3/57.8 \\
    NLI\textsuperscript{50\%} & 53.3/57.7 & 48.8/55.0 & 53.9/60.0 & 51.6/57.1 \\
    NLI\textsuperscript{75\%} & 54.1/\uline{60.2} & 50.7/55.2 & 51.7/59.1 & 52.2/58.0 \\
    \hline
    CLIP\textsuperscript{25\%} & 52.8/58.4 & 50.0/53.8 & 56.1/58.7 & 52.4/56.6 \\
    CLIP\textsuperscript{50\%} & 53.8/59.2 & 48.6/53.6 & 57.8/57.8 & 52.6/56.6 \\
    CLIP\textsuperscript{75\%} & 55.1/58.7 & 49.3/\uline{56.0} & 54.3/58.7 & 52.6/57.6 \\
    \hline
    NCLIP\textsuperscript{25\%} & 54.3/58.7 & 51.4/53.6 & 50.0/60.4 & 52.2/57.0 \\
    NCLIP\textsuperscript{50\%} & 54.6/59.9 & 50.0/54.0 & \textbf{58.8}/59.6 & 53.6/57.5 \\
    NCLIP\textsuperscript{75\%} & \textbf{57.1}/60.0 & 50.0/54.8 & \uline{58.3}/\uline{61.3} & \textbf{54.5}/\uline{58.2} \\
    \hline
    \rowcolor{rowgray}
    HiCQA\textsuperscript{25\%} & 53.8/57.7 & 51.0/53.3 & 57.0/58.3 & 53.4/56.0 \\
    \rowcolor{rowgray}
    HiCQA\textsuperscript{50\%} & 54.3/\uline{60.2} & 48.8/53.8 & 56.1/57.8 & 52.5/57.1 \\
    \rowcolor{rowgray}
    HiCQA\textsuperscript{75\%} & 55.4/\textbf{61.0} & \textbf{52.4}/\textbf{56.9} & 57.0/\uline{61.3} & \textbf{54.5}/\textbf{59.4} \\
    
    \Xhline{1.2pt}
    \end{tabular}
    \vspace{-8pt}
    \label{tab4}
\end{table}
\subsection{Ablation experiments}
We conduct three sets of ablations: \textbf{(i)} the effect of different weak supervised signals (w/o CLIP, w/o NLI and w/o Capa); \textbf{(ii)} the contribution of an additional cross-modal consistency token (w/o token); and \textbf{(iii)} training strategies (SFT with free-form QA, SFT with MCQ, GRPO).


\paragraph{Supervised signals.} Table~\ref{tab7-ab} shows the effect of different supervision signals. Similar to the results in Table~\ref{tab4}, removing any single weak supervision signal leads to a degradation in Avg performance. Removing the capacity supervision signal leads to a modest performance drop.


\paragraph{Cross-modal consistency.} As shown in Table~\ref{tab7-ab}. We append an extra similarity token to the embeddings of Image and QA nodes to encourage cross-modal consistency learning. Removing this token (HiCQA w/o t) degrades the average score, confirming its utility.


\paragraph{Training methods.} In Table~\ref{tab5}, SFT with MCQ delivers a large gain over SFT with QA, and GRPO with MCQ attains competitive performance as well, indicating that MCQ-format supervision is an effective learning signal for microscopy reasoning. Training GRPO on QA data without a format reward led to severe reward hacking~\cite{skalse2022rh2, yu2024rh1}, more details will be discussed in the appendix. 


\begin{table}[t]
    \centering
    \small
    \caption{Ablation experiments on HiCQA-Graph (2B/4B). }
    \vspace{-4pt}
    \begin{tabular}{lcccc}
    \Xhline{1.2pt}
    Methods & EU & HG & EP & Avg  \\
    \Xhline{0.5pt}
    \addlinespace[1.2pt]
    \Xhline{0.5pt}
    
    Ours\textsuperscript{w/o CLIP} & \textbf{56.4}/58.9 & 50.2/\textbf{56.9} & \textbf{57.4}/\textbf{61.3} & 54.1/\uline{58.6} \\
    Ours\textsuperscript{w/o NLI} & 54.1/59.7 & 50.5/\uline{56.2} & 56.5/\uline{60.0} & 53.2/58.3 \\
    Ours\textsuperscript{w/o token} & \uline{55.9}/\textbf{61.7} & \uline{51.9}/53.3 & 55.7/57.4 & \uline{54.2}/57.4 \\
    Ours\textsuperscript{w/o Capa} & 55.1/59.7 & \uline{51.9}/55.0 & 56.1/60.4 & 54.0/58.0 \\
    Ours\textsuperscript{best} & 55.4/\uline{61.0} & \textbf{52.4}/\textbf{56.9} & \uline{57.0}/\textbf{61.3} & \textbf{54.5}/\textbf{59.4} \\
    \Xhline{1.2pt}
    \end{tabular}
    
    \label{tab7-ab}
    \vspace{-4pt}
\end{table}

\begin{table}[t]
    \centering
    \small
    \caption{Ablation experiments on training methods.}
    \begin{tabular}{lcccc}
    \Xhline{1.2pt}
    Models & EU & HG & EP & Avg \\
    \Xhline{0.5pt}
    \addlinespace[1.2pt]
    \Xhline{0.5pt}
    SFT w/ QA & 49.2 & 39.5 & 36.1 & 42.4 \\
    SFT w/ MCQ & \textbf{61.0} & \textbf{56.9} & \textbf{61.3} & \textbf{59.4} \\
    GRPO w/ QA & - & - & - & - \\
    GRPO w/ MCQ & \uline{58.7} & \uline{54.5} & \uline{60.4} & \uline{57.4} \\
    \Xhline{1.2pt}
    \end{tabular}
    
    \label{tab5}
\end{table}


\begin{table}[htbp]
    \centering 
    \small
    \caption{Computational overheads (ms) of HiCQA per image.} 
    \vspace{-6pt}
    \label{tab6}
    \begin{tabular}{lccc}
    \Xhline{1.2pt}
    Component & forward & backward & end-to-end \\
    \Xhline{0.5pt}
    \addlinespace[1.2pt]
    \Xhline{0.5pt}
    CLIP & 32.29 & - & - \\
    NLI & 77.05 & - & - \\
    Train & 3.56 & 10.21 & 129.32 \\
    Test & 0.63 & - & 109.82 \\
    \Xhline{1.2pt}
\end{tabular}
\vspace{-5pt}
\end{table}

\subsection{Error analysis on MicroVQA++}
We briefly summarize typical failure modes observed on the MicroVQA++ test set (case studies are deferred to appendix). \textbf{(i) Visual grounding}: mislocalizing the referenced structure or channel, especially for small or overlapping compartments, leading to plausible but wrong EU answers. \textbf{(ii) Distractor susceptibility}: over-reliance on lexical overlap yields selection of semantically close but incorrect MCQ options, indicating shortcut pattern matching. \textbf{(iii) Cross-modal priors}: caption-like hallucinations and overuse of familiar biomedical tropes (e.g., infection/inflammation) that are unsupported by the image.

\subsection{Computational overheads}
We report per-image latency for each stage on a single RTX 3090 GPU (Table~\ref{tab6}). Graph construction is dominated by NLI inference and CLIP encoding, while graph training and inference are lightweight. The end-to-end training and inference times are 129.42 ms and 109.82 ms, respectively.

%% file: sec/5_conclusion.tex
\section{Conclusion}

We present \emph{MicroVQA++}, a microscopy-focused VQA dataset generation pipeline that converts expert images and captions in PubMed articles into QA and MCQ via a MLLM agent and enforces cross-modal consistency with a HiCQA-Graph before human checks, substantially improving supervision. The resulting \emph{MicroVQA++} test set is larger and harder in Bloom levels, providing a stricter evaluation. After closed-form MCQs SFT on \emph{MicroVQA++} train set, a small InternVL3.5-4B-Instruct rivals top proprietary systems on \emph{MicroVQA} and sets new SOTA among open-source MLLMs; In general, carefully building high-quality dataset enables small MLLMs to advance in microscopy reasoning. Limitations and future work available in appendix.

%% file: main.bbl
\begin{thebibliography}{56}
\providecommand{\natexlab}[1]{#1}
\providecommand{\url}[1]{\texttt{#1}}
\expandafter\ifx\csname urlstyle\endcsname\relax
  \providecommand{\doi}[1]{doi: #1}\else
  \providecommand{\doi}{doi: \begingroup \urlstyle{rm}\Url}\fi

\bibitem[Ahn and Kwak(2018)]{ahn2018g3}
Jiwoon Ahn and Suha Kwak.
\newblock Learning pixel-level semantic affinity with image-level supervision for weakly supervised semantic segmentation.
\newblock In \emph{Proceedings of the IEEE conference on computer vision and pattern recognition}, pages 4981--4990, 2018.

\bibitem[Alhazmi et~al.(2024)Alhazmi, Sheng, Zhang, Zaib, and Alhazmi]{alhazmi2024mcq2}
Elaf Alhazmi, Quan~Z Sheng, Wei~Emma Zhang, Munazza Zaib, and Ahoud Alhazmi.
\newblock Distractor generation for multiple-choice questions: A survey of methods, datasets, and evaluation.
\newblock \emph{arXiv preprint arXiv:2402.01512}, 3\penalty0 (5), 2024.

\bibitem[{Anthropic}(2025)]{anthropic_claude_sonnet_4_5_system_card_2025}
{Anthropic}.
\newblock Claude sonnet 4.5 system card, 2025.

\bibitem[Baghbanzadeh et~al.(2025)Baghbanzadeh, Ashkezari, Dolatabadi, and Afkanpour]{baghbanzadeh2025moti1}
Negin Baghbanzadeh, Sajad Ashkezari, Elham Dolatabadi, and Arash Afkanpour.
\newblock Open-pmc-18m: A high-fidelity large scale medical dataset for multimodal representation learning.
\newblock \emph{arXiv preprint arXiv:2506.02738}, 2025.

\bibitem[Bai et~al.(2023)Bai, Bai, Chu, Cui, Dang, Deng, Fan, Ge, Han, Huang, et~al.]{bai2023qwen}
Jinze Bai, Shuai Bai, Yunfei Chu, Zeyu Cui, Kai Dang, Xiaodong Deng, Yang Fan, Wenbin Ge, Yu Han, Fei Huang, et~al.
\newblock Qwen technical report.
\newblock \emph{arXiv preprint arXiv:2309.16609}, 2023.

\bibitem[Bai et~al.(2025)Bai, Chen, Liu, Wang, Ge, Song, Dang, Wang, Wang, Tang, et~al.]{bai2025qwen25}
Shuai Bai, Keqin Chen, Xuejing Liu, Jialin Wang, Wenbin Ge, Sibo Song, Kai Dang, Peng Wang, Shijie Wang, Jun Tang, et~al.
\newblock Qwen2. 5-vl technical report.
\newblock \emph{arXiv preprint arXiv:2502.13923}, 2025.

\bibitem[Bitew et~al.(2023)Bitew, Deleu, Develder, and Demeester]{bitew2023qagen1}
Semere~Kiros Bitew, Johannes Deleu, Chris Develder, and Thomas Demeester.
\newblock Distractor generation for multiple-choice questions with predictive prompting and large language models.
\newblock In \emph{Joint European Conference on Machine Learning and Knowledge Discovery in Databases}, pages 48--63. Springer, 2023.

\bibitem[Bowman et~al.(2015)Bowman, Angeli, Potts, and Manning]{bowman2015large}
Samuel Bowman, Gabor Angeli, Christopher Potts, and Christopher~D Manning.
\newblock A large annotated corpus for learning natural language inference.
\newblock In \emph{Proceedings of the 2015 conference on empirical methods in natural language processing}, pages 632--642, 2015.

\bibitem[Boykov and Jolly(2001)]{boykov2001g1}
Yuri~Y Boykov and M-P Jolly.
\newblock Interactive graph cuts for optimal boundary \& region segmentation of objects in nd images.
\newblock In \emph{Proceedings eighth IEEE international conference on computer vision. ICCV 2001}, pages 105--112. IEEE, 2001.

\bibitem[Brody et~al.(2021)Brody, Alon, and Yahav]{brody2021gatv2}
Shaked Brody, Uri Alon, and Eran Yahav.
\newblock How attentive are graph attention networks?
\newblock \emph{arXiv preprint arXiv:2105.14491}, 2021.

\bibitem[Burgess et~al.(2025)Burgess, Nirschl, Bravo-S{\'a}nchez, Lozano, Gupte, Galaz-Montoya, Zhang, Su, Bhowmik, Coman, et~al.]{burgess2025microvqa}
James Burgess, Jeffrey~J Nirschl, Laura Bravo-S{\'a}nchez, Alejandro Lozano, Sanket~Rajan Gupte, Jesus~G Galaz-Montoya, Yuhui Zhang, Yuchang Su, Disha Bhowmik, Zachary Coman, et~al.
\newblock Microvqa: A multimodal reasoning benchmark for microscopy-based scientific research.
\newblock In \emph{Proceedings of the Computer Vision and Pattern Recognition Conference}, pages 19552--19564, 2025.

\bibitem[Choudhary(2025)]{choudhary2025microscopygpt}
Kamal Choudhary.
\newblock Microscopygpt: Generating atomic-structure captions from microscopy images of 2d materials with vision-language transformers.
\newblock \emph{The Journal of Physical Chemistry Letters}, 16:\penalty0 7028--7035, 2025.

\bibitem[Conneau et~al.(2018)Conneau, Rinott, Lample, Williams, Bowman, Schwenk, and Stoyanov]{conneau2018xnli}
Alexis Conneau, Ruty Rinott, Guillaume Lample, Adina Williams, Samuel Bowman, Holger Schwenk, and Veselin Stoyanov.
\newblock Xnli: Evaluating cross-lingual sentence representations.
\newblock In \emph{Proceedings of the 2018 conference on empirical methods in natural language processing}, pages 2475--2485, 2018.

\bibitem[Conneau et~al.(2020)Conneau, Khandelwal, Goyal, Chaudhary, Wenzek, Guzm{\'a}n, Grave, Ott, Zettlemoyer, and Stoyanov]{conneau2020xlmr}
Alexis Conneau, Kartikay Khandelwal, Naman Goyal, Vishrav Chaudhary, Guillaume Wenzek, Francisco Guzm{\'a}n, Edouard Grave, Myle Ott, Luke Zettlemoyer, and Veselin Stoyanov.
\newblock Unsupervised cross-lingual representation learning at scale.
\newblock In \emph{Proceedings of the 58th annual meeting of the association for computational linguistics}, pages 8440--8451, 2020.

\bibitem[Dosovitskiy(2020)]{dosovitskiy2020vit}
Alexey Dosovitskiy.
\newblock An image is worth 16x16 words: Transformers for image recognition at scale.
\newblock \emph{arXiv preprint arXiv:2010.11929}, 2020.

\bibitem[Fey and Lenssen(2019)]{fey2019fast}
Matthias Fey and Jan~Eric Lenssen.
\newblock Fast graph representation learning with pytorch geometric.
\newblock \emph{arXiv preprint arXiv:1903.02428}, 2019.

\bibitem[Gierl et~al.(2017)Gierl, Bulut, Guo, and Zhang]{gierl2017mcq1}
Mark~J Gierl, Okan Bulut, Qi Guo, and Xinxin Zhang.
\newblock Developing, analyzing, and using distractors for multiple-choice tests in education: A comprehensive review.
\newblock \emph{Review of educational research}, 87\penalty0 (6):\penalty0 1082--1116, 2017.

\bibitem[Goals(1956)]{goals1956bloom}
Educational Goals.
\newblock Handbook i: Cognitive domain.
\newblock \emph{New York: David}, 1956.

\bibitem[Hamilton et~al.(2017)Hamilton, Ying, and Leskovec]{hamilton2017graphsage}
Will Hamilton, Zhitao Ying, and Jure Leskovec.
\newblock Inductive representation learning on large graphs.
\newblock \emph{Advances in neural information processing systems}, 30, 2017.

\bibitem[Hu et~al.(2022)Hu, Shen, Wallis, Allen-Zhu, Li, Wang, Wang, Chen, et~al.]{hu2022lora}
Edward~J Hu, Yelong Shen, Phillip Wallis, Zeyuan Allen-Zhu, Yuanzhi Li, Shean Wang, Lu Wang, Weizhu Chen, et~al.
\newblock Lora: Low-rank adaptation of large language models.
\newblock \emph{ICLR}, 1\penalty0 (2):\penalty0 3, 2022.

\bibitem[Hurst et~al.(2024)Hurst, Lerer, Goucher, Perelman, Ramesh, Clark, Ostrow, Welihinda, Hayes, Radford, et~al.]{openai_gpt4o}
Aaron Hurst, Adam Lerer, Adam~P Goucher, Adam Perelman, Aditya Ramesh, Aidan Clark, AJ Ostrow, Akila Welihinda, Alan Hayes, Alec Radford, et~al.
\newblock Gpt-4o system card.
\newblock \emph{arXiv preprint arXiv:2410.21276}, 2024.

\bibitem[Ilharco et~al.(2021)Ilharco, Wortsman, Wightman, Gordon, Carlini, Taori, Dave, Shankar, Namkoong, Miller, Hajishirzi, Farhadi, and Schmidt]{openclip}
Gabriel Ilharco, Mitchell Wortsman, Ross Wightman, Cade Gordon, Nicholas Carlini, Rohan Taori, Achal Dave, Vaishaal Shankar, Hongseok Namkoong, John Miller, Hannaneh Hajishirzi, Ali Farhadi, and Ludwig Schmidt.
\newblock Openclip, 2021.

\bibitem[Jaech et~al.(2024)Jaech, Kalai, Lerer, Richardson, El-Kishky, Low, Helyar, Madry, Beutel, Carney, et~al.]{openai_o1_series}
Aaron Jaech, Adam Kalai, Adam Lerer, Adam Richardson, Ahmed El-Kishky, Aiden Low, Alec Helyar, Aleksander Madry, Alex Beutel, Alex Carney, et~al.
\newblock Openai o1 system card.
\newblock \emph{arXiv preprint arXiv:2412.16720}, 2024.

\bibitem[Jiang et~al.(2024)Jiang, Li, Deng, Liu, Gao, Zhou, Li, Wang, and Zheng]{jiang2024anomaly}
Xi Jiang, Jian Li, Hanqiu Deng, Yong Liu, Bin-Bin Gao, Yifeng Zhou, Jialin Li, Chengjie Wang, and Feng Zheng.
\newblock Mmad: A comprehensive benchmark for multimodal large language models in industrial anomaly detection.
\newblock \emph{arXiv preprint arXiv:2410.09453}, 2024.

\bibitem[Kr{\"a}henb{\"u}hl and Koltun(2011)]{krahenbuhl2011g2}
Philipp Kr{\"a}henb{\"u}hl and Vladlen Koltun.
\newblock Efficient inference in fully connected crfs with gaussian edge potentials.
\newblock \emph{Advances in neural information processing systems}, 24, 2011.

\bibitem[Li et~al.(2023)Li, Wong, Zhang, Usuyama, Liu, Yang, Naumann, Poon, and Gao]{li2023llavamed}
Chunyuan Li, Cliff Wong, Sheng Zhang, Naoto Usuyama, Haotian Liu, Jianwei Yang, Tristan Naumann, Hoifung Poon, and Jianfeng Gao.
\newblock Llava-med: Training a large language-and-vision assistant for biomedicine in one day.
\newblock \emph{Advances in Neural Information Processing Systems}, 36:\penalty0 28541--28564, 2023.

\bibitem[Lin et~al.(2023)Lin, Zhao, Zhang, Wu, Zhang, Wang, and Xie]{lin2023moti3}
Weixiong Lin, Ziheng Zhao, Xiaoman Zhang, Chaoyi Wu, Ya Zhang, Yanfeng Wang, and Weidi Xie.
\newblock Pmc-clip: Contrastive language-image pre-training using biomedical documents.
\newblock In \emph{International Conference on Medical Image Computing and Computer-Assisted Intervention}, pages 525--536. Springer, 2023.

\bibitem[Liu et~al.(2019)Liu, Ott, Goyal, Du, Joshi, Chen, Levy, Lewis, Zettlemoyer, and Stoyanov]{liu2019roberta}
Yinhan Liu, Myle Ott, Naman Goyal, Jingfei Du, Mandar Joshi, Danqi Chen, Omer Levy, Mike Lewis, Luke Zettlemoyer, and Veselin Stoyanov.
\newblock Roberta: A robustly optimized bert pretraining approach.
\newblock \emph{arXiv preprint arXiv:1907.11692}, 2019.

\bibitem[Lozano et~al.(2024)Lozano, Nirschl, Burgess, Gupte, Zhang, Unell, and Yeung]{lozano2024ubench}
Alejandro Lozano, Jeffrey Nirschl, James Burgess, Sanket~Rajan Gupte, Yuhui Zhang, Alyssa Unell, and Serena Yeung.
\newblock Micro-bench: A microscopy benchmark for vision-language understanding.
\newblock \emph{Advances in Neural Information Processing Systems}, 37:\penalty0 30670--30685, 2024.

\bibitem[Lozano et~al.(2025)Lozano, Sun, Burgess, Chen, Nirschl, Gu, Lopez, Aklilu, Rau, Katzer, et~al.]{lozano2025biomedica}
Alejandro Lozano, Min~Woo Sun, James Burgess, Liangyu Chen, Jeffrey~J Nirschl, Jeffrey Gu, Ivan Lopez, Josiah Aklilu, Anita Rau, Austin~Wolfgang Katzer, et~al.
\newblock Biomedica: An open biomedical image-caption archive, dataset, and vision-language models derived from scientific literature.
\newblock In \emph{Proceedings of the Computer Vision and Pattern Recognition Conference}, pages 19724--19735, 2025.

\bibitem[Maaten and Hinton(2008)]{maaten2008visualizing}
Laurens van~der Maaten and Geoffrey Hinton.
\newblock Visualizing data using t-sne.
\newblock \emph{Journal of machine learning research}, 9\penalty0 (Nov):\penalty0 2579--2605, 2008.

\bibitem[Mandal et~al.(2025)Mandal, Soni, Zaki, Smedskjaer, Wondraczek, Wondraczek, Gosvami, and Krishnan]{mandal2025evaluating}
Indrajeet Mandal, Jitendra Soni, Mohd Zaki, Morten~M Smedskjaer, Katrin Wondraczek, Lothar Wondraczek, Nitya~Nand Gosvami, and NM~Anoop Krishnan.
\newblock Evaluating large language model agents for automation of atomic force microscopy.
\newblock \emph{Nature Communications}, 16\penalty0 (1):\penalty0 9104, 2025.

\bibitem[Nam et~al.(2025)Nam, Kim, Kyung, Seo, Song, Kwon, Kim, Jo, Park, Sung, et~al.]{nam2025medicalimaging}
Yoojin Nam, Dong~Yeong Kim, Sunggu Kyung, Jinyoung Seo, Jeong~Min Song, Jimin Kwon, Jihyun Kim, Wooyoung Jo, Hyungbin Park, Jimin Sung, et~al.
\newblock Multimodal large language models in medical imaging: Current state and future directions.
\newblock \emph{Korean Journal of Radiology}, 26\penalty0 (10):\penalty0 900, 2025.

\bibitem[{OpenAI}(2025{\natexlab{a}})]{openai_gpt5_system_card_2025}
{OpenAI}.
\newblock Gpt-5 system card, 2025{\natexlab{a}}.

\bibitem[{OpenAI}(2025{\natexlab{b}})]{openai_o3_o4mini_systemcard_2025}
{OpenAI}.
\newblock Openai o3 and o4-mini system card, 2025{\natexlab{b}}.

\bibitem[Radford et~al.(2021)Radford, Kim, Hallacy, Ramesh, Goh, Agarwal, Sastry, Askell, Mishkin, Clark, et~al.]{radford2021clip}
Alec Radford, Jong~Wook Kim, Chris Hallacy, Aditya Ramesh, Gabriel Goh, Sandhini Agarwal, Girish Sastry, Amanda Askell, Pamela Mishkin, Jack Clark, et~al.
\newblock Learning transferable visual models from natural language supervision.
\newblock In \emph{International conference on machine learning}, pages 8748--8763. PmLR, 2021.

\bibitem[Ratner et~al.(2016)Ratner, De~Sa, Wu, Selsam, and R{\'e}]{ratner2016relatedgraph1}
Alexander~J Ratner, Christopher~M De~Sa, Sen Wu, Daniel Selsam, and Christopher R{\'e}.
\newblock Data programming: Creating large training sets, quickly.
\newblock \emph{Advances in neural information processing systems}, 29, 2016.

\bibitem[R{\"u}ckert et~al.(2024)R{\"u}ckert, Bloch, Br{\"u}ngel, Idrissi-Yaghir, Sch{\"a}fer, Schmidt, Koitka, Pelka, Abacha, G.~Seco~de Herrera, et~al.]{ruckert2024moti2}
Johannes R{\"u}ckert, Louise Bloch, Raphael Br{\"u}ngel, Ahmad Idrissi-Yaghir, Henning Sch{\"a}fer, Cynthia~S Schmidt, Sven Koitka, Obioma Pelka, Asma~Ben Abacha, Alba G.~Seco~de Herrera, et~al.
\newblock Rocov2: Radiology objects in context version 2, an updated multimodal image dataset.
\newblock \emph{Scientific Data}, 11\penalty0 (1):\penalty0 688, 2024.

\bibitem[Shao et~al.(2024)Shao, Wang, Zhu, Xu, Song, Bi, Zhang, Zhang, Li, Wu, et~al.]{grpo}
Zhihong Shao, Peiyi Wang, Qihao Zhu, Runxin Xu, Junxiao Song, Xiao Bi, Haowei Zhang, Mingchuan Zhang, YK Li, Yang Wu, et~al.
\newblock Deepseekmath: Pushing the limits of mathematical reasoning in open language models.
\newblock \emph{arXiv preprint arXiv:2402.03300}, 2024.

\bibitem[Sheng et~al.(2025)Sheng, Zhang, Ye, Wu, Zhang, Zhang, Peng, Lin, and Wu]{sheng2025verl}
Guangming Sheng, Chi Zhang, Zilingfeng Ye, Xibin Wu, Wang Zhang, Ru Zhang, Yanghua Peng, Haibin Lin, and Chuan Wu.
\newblock Hybridflow: A flexible and efficient rlhf framework.
\newblock In \emph{Proceedings of the Twentieth European Conference on Computer Systems}, pages 1279--1297, 2025.

\bibitem[Skalse et~al.(2022)Skalse, Howe, Krasheninnikov, and Krueger]{skalse2022rh2}
Joar Skalse, Nikolaus Howe, Dmitrii Krasheninnikov, and David Krueger.
\newblock Defining and characterizing reward gaming.
\newblock \emph{Advances in Neural Information Processing Systems}, 35:\penalty0 9460--9471, 2022.

\bibitem[Vaswani et~al.(2017)Vaswani, Shazeer, Parmar, Uszkoreit, Jones, Gomez, Kaiser, and Polosukhin]{vaswani2017transformer}
Ashish Vaswani, Noam Shazeer, Niki Parmar, Jakob Uszkoreit, Llion Jones, Aidan~N Gomez, {\L}ukasz Kaiser, and Illia Polosukhin.
\newblock Attention is all you need.
\newblock \emph{Advances in neural information processing systems}, 30, 2017.

\bibitem[Velickovic et~al.(2017)Velickovic, Cucurull, Casanova, Romero, Lio, Bengio, et~al.]{velickovic2017gat}
Petar Velickovic, Guillem Cucurull, Arantxa Casanova, Adriana Romero, Pietro Lio, Yoshua Bengio, et~al.
\newblock Graph attention networks.
\newblock \emph{stat}, 1050\penalty0 (20):\penalty0 10--48550, 2017.

\bibitem[Wang et~al.(2024)Wang, Bai, Tan, Wang, Fan, Bai, Chen, Liu, Wang, Ge, et~al.]{wang2024qwen2vl}
Peng Wang, Shuai Bai, Sinan Tan, Shijie Wang, Zhihao Fan, Jinze Bai, Keqin Chen, Xuejing Liu, Jialin Wang, Wenbin Ge, et~al.
\newblock Qwen2-vl: Enhancing vision-language model's perception of the world at any resolution.
\newblock \emph{arXiv preprint arXiv:2409.12191}, 2024.

\bibitem[Wang et~al.(2025)Wang, Gao, Gu, Pu, Cui, Wei, Liu, Jing, Ye, Shao, et~al.]{wang2025internvl35}
Weiyun Wang, Zhangwei Gao, Lixin Gu, Hengjun Pu, Long Cui, Xingguang Wei, Zhaoyang Liu, Linglin Jing, Shenglong Ye, Jie Shao, et~al.
\newblock Internvl3. 5: Advancing open-source multimodal models in versatility, reasoning, and efficiency.
\newblock \emph{arXiv preprint arXiv:2508.18265}, 2025.

\bibitem[Wei et~al.(2022)Wei, Wang, Schuurmans, Bosma, Xia, Chi, Le, Zhou, et~al.]{wei2022cot}
Jason Wei, Xuezhi Wang, Dale Schuurmans, Maarten Bosma, Fei Xia, Ed Chi, Quoc~V Le, Denny Zhou, et~al.
\newblock Chain-of-thought prompting elicits reasoning in large language models.
\newblock \emph{Advances in neural information processing systems}, 35:\penalty0 24824--24837, 2022.

\bibitem[Williams et~al.(2018)Williams, Nangia, and Bowman]{williams2018mnli}
Adina Williams, Nikita Nangia, and Samuel Bowman.
\newblock A broad-coverage challenge corpus for sentence understanding through inference.
\newblock In \emph{Proceedings of the 2018 conference of the North American chapter of the association for computational linguistics: human language technologies, volume 1 (long papers)}, pages 1112--1122, 2018.

\bibitem[Wu et~al.(2021)Wu, Wei, Jiang, Mao, Tang, and Li]{wu2021ngc}
Zhi-Fan Wu, Tong Wei, Jianwen Jiang, Chaojie Mao, Mingqian Tang, and Yu-Feng Li.
\newblock Ngc: A unified framework for learning with open-world noisy data.
\newblock In \emph{Proceedings of the IEEE/CVF International Conference on Computer Vision}, pages 62--71, 2021.

\bibitem[Yang et~al.(2025)Yang, Li, Yang, Zhang, Hui, Zheng, Yu, Gao, Huang, Lv, et~al.]{yang2025qwen3}
An Yang, Anfeng Li, Baosong Yang, Beichen Zhang, Binyuan Hui, Bo Zheng, Bowen Yu, Chang Gao, Chengen Huang, Chenxu Lv, et~al.
\newblock Qwen3 technical report.
\newblock \emph{arXiv preprint arXiv:2505.09388}, 2025.

\bibitem[Yu et~al.(2024)Yu, Yao, Zhang, He, Han, Cui, Hu, Liu, Zheng, Sun, et~al.]{yu2024rh1}
Tianyu Yu, Yuan Yao, Haoye Zhang, Taiwen He, Yifeng Han, Ganqu Cui, Jinyi Hu, Zhiyuan Liu, Hai-Tao Zheng, Maosong Sun, et~al.
\newblock Rlhf-v: Towards trustworthy mllms via behavior alignment from fine-grained correctional human feedback.
\newblock In \emph{Proceedings of the IEEE/CVF Conference on Computer Vision and Pattern Recognition}, pages 13807--13816, 2024.

\bibitem[Zhang et~al.(2021)Zhang, Xing, and Liu]{zhang2021dualgraph}
HaiYang Zhang, XiMing Xing, and Liang Liu.
\newblock Dualgraph: A graph-based method for reasoning about label noise.
\newblock In \emph{Proceedings of the IEEE/CVF conference on computer vision and pattern recognition}, pages 9654--9663, 2021.

\bibitem[Zhang et~al.(2023)Zhang, Xu, Usuyama, Bagga, Tinn, Preston, Rao, Wei, Valluri, Wong, et~al.]{zhang2023moti4}
Sheng Zhang, Yanbo Xu, Naoto Usuyama, Jaspreet Bagga, Robert Tinn, Sam Preston, Rajesh Rao, Mu Wei, Naveen Valluri, Cliff Wong, et~al.
\newblock Large-scale domain-specific pretraining for biomedical vision-language processing.
\newblock \emph{arXiv preprint arXiv:2303.00915}, 2\penalty0 (3):\penalty0 6, 2023.

\bibitem[Zheng et~al.(2024)Zheng, Zhang, Zhang, Ye, Luo, Feng, and Ma]{zheng2024llamafactory}
Yaowei Zheng, Richong Zhang, Junhao Zhang, Yanhan Ye, Zheyan Luo, Zhangchi Feng, and Yongqiang Ma.
\newblock Llamafactory: Unified efficient fine-tuning of 100+ language models.
\newblock \emph{arXiv preprint arXiv:2403.13372}, 2024.

\bibitem[Zhou et~al.(2025)Zhou, Qian, and Gamba]{zhou2025remote}
Guoqing Zhou, Lihuang Qian, and Paolo Gamba.
\newblock Advances on multimodal remote sensing foundation models for earth observation downstream tasks: A survey.
\newblock \emph{Remote Sensing}, 2025.

\bibitem[Zhou et~al.(2024)Zhou, Liu, Yurtsever, Zagar, Zimmer, Cao, and Knoll]{zhou2024autonomous}
Xingcheng Zhou, Mingyu Liu, Ekim Yurtsever, Bare~Luka Zagar, Walter Zimmer, Hu Cao, and Alois~C Knoll.
\newblock Vision language models in autonomous driving: A survey and outlook.
\newblock \emph{IEEE Transactions on Intelligent Vehicles}, 2024.

\bibitem[Zhu and Ghahramani(2002)]{zhu2002relatedgraph2}
Xiaojin Zhu and Zoubin Ghahramani.
\newblock Learning from labeled and unlabeled data with label propagation.
\newblock \emph{ProQuest number: information to all users}, 2002.

\end{thebibliography}
